# Learning More with Less: GAN-based Medical Image Augmentation

Changhee HAN[*1*2], Kohei MURAO[*2], Shin'ichi SATOH[*2], Hideki NAKAYAMA[*1]

**Abstract**

Convolutional Neural Network (CNN)-based accurate prediction typically requires large-scale annotated training data. In Medical Imaging, however, both obtaining medical data and annotating them by expert physicians are challenging; to overcome this lack of data, Data Augmentation (DA) using Generative Adversarial Networks (GANs) is essential, since they can synthesize additional annotated training data to handle small and fragmented medical images from various scanners—those generated images, realistic but completely novel, can further fill the real image distribution uncovered by the original dataset. As a tutorial, this paper introduces GAN-based Medical Image Augmentation, along with tricks to boost classification/object detection/segmentation performance using them, based on our experience and related work. Moreover, we show our first GAN-based DA work using automatic bounding box annotation, for robust CNN-based brain metastases detection on 256 × 256 MR images; GAN-based DA can boost 10% sensitivity in diagnosis with a clinically acceptable number of additional False Positives, even with highly-rough and inconsistent bounding boxes.

**Keywords** : Generative Adversarial Networks, Data Augmentation, Medical Image Augmentation

## 1. Introduction

Convolutional Neural Networks (CNNs) have shown state-of-the-art performance in various applications, primarily thanks to large-scale annotated training data. Unfortunately, obtaining such huge annotated medical images is challenging; thus, as classical Data Augmentation (DA) techniques, we generally use geometric/intensity transformations of original images for accurate diagnosis. Those transformed/original images, however, intrinsically have similar distributions, causing partially improved performance; in this regard, Generative Adversarial Network (GAN) [1]-based DA can further boost the performance by filling the uncovered real image distribution, since it synthesizes realistic but completely novel samples with a good generalization ability; in Computer Vision, GANs showed remarkable DA performance, such as SimGAN's 21% performance improvement in eye-gaze estimation [2]. This GAN-based DA trend especially applies to Medical Imaging for handling various types of small and fragmented medical datasets from multiple scanners: some researchers used it for classification on brain tumor MR [3] and skin lesion images [4]; others used it for segmentation on 3D lung nodule CT images [5].

## 2. Tricks to Learn More with Less

1) Pre-processing

―――――――――――――――――――――――――――――――――――――――――――――――

*1 Graduate School of Information Science and Technology, The University of Tokyo,
  〔Yayoi 1-1-1, Bunkyo, Tokyo, 113-8657〕
  e-mail: han@nlab.ci.i.u-tokyo.ac.jp

*2 Research Center for Medical Bigdata, National Institute of Informatics.

Since variety in size, location, shape, and visual appearance heavily influences GAN performance especially when training data are small and fragmented, generating desired images requires 2-step pre-processing: (*i*) Removing irrelevant information on medical images (e.g., denoising/skull-stripping); (*ii*) Cropping and resizing the remaining parts to a power of 2 (e.g., $256 \times 256/32 \times 32 \times 32$). The Spatial Pyramid Pooling (SPP) layer in the discriminator can be used to avoid the effect of resizing [6]. Classical DA techniques (i.e., geometric/intensity transformations) may help GAN training (at least horizontal and vertical flipping), along with Webly Supervised Learning [7] for external data.

**2) GAN-based Image Generation**

We can either generate whole images including pathological parts (e.g., whole brain MR images with tumors) [1] or Regions of Interest (ROIs) alone (e.g., skin lesion images) [4] for 2D image generation, but only ROIs alone (e.g., 3D lung CT nodules) [5] for 3D image generation due to heavy computation power—if needed, we can paste the generated ROIs naturally to the whole images for DA. Moreover, Progressive Growing of GANs (PGGANs) [8] with the Wasserstein loss using gradient penalty [9] can generate a variety of realistic high-resolution (i.e., $\geq 256 \times 256$) medical images. From the latent space, whether to sample points using a normal or uniform distribution does not empirically make a significant difference to DA.

**3) Post-processing**

GAN-generated samples often contain visible artifacts; but, discarding such images with weird artifacts does not boost DA, since realism confirmed by humans—including *via* Visual Turing Test [10] by expert physicians [11]—is not strongly associated with DA performance. Therefore, use both success/failure cases for DA without post-processing, unless the synthetic images clearly lack disease appearance.

**4) Image-Quality Evaluation on GAN-generated Images**

T-Distributed Stochastic Neighbor Embedding (t-SNE) algorithm [12] can visualize the distribution of real and synthetic images by directly embedding those high-dimensional data into a 2D/3D space. The Visual Turing Test by physicians can also quantitatively evaluate how realistic or certain disease-like the GAN-based synthetic images are by supplying, in a random order, a random selection of the same number of real/synthetic images [11] per each class.

**5) Application to Classification**

We can use GAN-generated images for classification, generating normal *vs* pathological images (e.g., non-tumor *vs* tumor) [3] or several types of pathological images [4]. Frid-Adar et al. reported better DA performance when generating labeled examples for each pathological class separately, rather than incorporating class conditioning to generate them all at once [4]. Balancing between real/synthetic images is essential during classifier training since adding over-sufficient synthetic images causes worse performance [3]; we can either classically augment only real or both real/synthetic images (empirically, the best balance between real/synthetic images is 1-to-1 or 1-to-2). Pre-training on ImageNet might not achieve better sensitivity than training from scratch. T-SNE can differentiate classification results with/without DA by visualizing features extracted from the last layer of each trained classifier.

**6) Application to Object Detection**

Moreover, to locate disease areas, we can exploit GANs for Object Detection; incorporating bounding box conditions into the GANs during training and generating images based on the annotation slightly different from training images during testing. Adding synthetic training images can achieve higher sensitivity with more False Positives (FPs). The next section describes our novel GAN-based DA work using bounding boxes [13].

**7) Application to Segmentation**

Similarly to Object Detection, we can condition rigorous segmentation into GANs to achieve higher Dice score.

## 3. GAN-based MR Image Augmentation for Brain Metastases Detection

**1) Brain Metastases Dataset**

As a small/fragmented dataset, we collected T1c brain axial MR images of 180 brain metastases cases from multiple MRI scanners. For tumor detection, the whole dataset (180 patients) is divided into: (*i*) A training set (126 patients/2,813 images/5,963 bounding boxes); (*ii*) A validation set (18 patients/337 images/616 bounding boxes); (*iii*) A test set (36 patients/947 images/3,094 bounding boxes); we only use the training set for the GAN training to be fair.

We first skull-strip all images with diverse resolution, and then crop the remaining brain parts and resize them to $256 \times 256$ pixels. In our first GAN-based medical DA for object detection, expert physicians lazily annotate tumors with highly-rough and inconsistent bounding boxes for labor minimization.

**2) Proposed GAN-based Image Generation**

As a preliminary study, we propose an innovative training method for GANs, incorporating bounding box conditions into PGGANs [8]. In the original PGGANs, a generator and discriminator are progressively growing: starting from low resolution, new layers model details as training progresses. We further condition the generator/discriminator to generate random, but realistic $256 \times 256$ MR images with tumors of random shape within bounding boxes. **Fig. 1** shows example real/GAN-generated images, including tumor bounding boxes.

**3) Brain Metastases Detection Using YOLOv3**

YOLOv3 [14], real-time and accurate CNN-based object detector, divides the image into regions to predict bounding boxes/probabilities for each region. We detect brain metastases using YOLOv3 for real-time tumor alert. To confirm the GAN-based DA influence, we compare detection results of 2,813 real images with/without 4,000 GAN-based DA. Not to overlook the diagnosis *via* computer-assisted diagnosis, higher sensitivity matters more than fewer FPs; thus, adding the GAN-generated training images, we try to obtain higher sensitivity with clinically acceptable FPs. Considering our highly-rough annotation, we calculate sensitivity/FPs per slice with both Intersection over Union (IoU) threshold 0.5 and 0.25. We also manually discard synthetic images with unclear tumor appearance for better DA.

**4) Brain Metastases Detection Results**

As shown in **Table 1**, the additional GAN-generated training images allow to remarkably increase the mean Average Precision (mAP) and sensitivity in return for more FPs per slice; the increased FPs derive from also detecting blood vessels, since they resemble the enhanced tumor regions small/hyper-intense on T1c MR images, due to the contrast agent perfused throughout the blood vessels. Additional 4,000 synthetic images improve sensitivity by 0.10 with IoU threshold 0.5 and by 0.08 with IoU threshold 0.25—they further fill the real image distribution and improve robustness during training, obtaining sensitivity 0.91 with moderate IoU threshold 0.25 even with the highly-rough bounding box annotation. **Fig. 2** also visually shows its ability to alleviate the risk of overlooking the tumor diagnosis, with clinically acceptable FPs (i.e., the bounding boxes highly-overlapping around tumors only require a physician's smooth single check by switching on/off transparent alpha-blended annotations on MR images in the clinical routine). It should be noted that we cannot increase FPs to achieve such high sensitivity without the GAN-based DA.

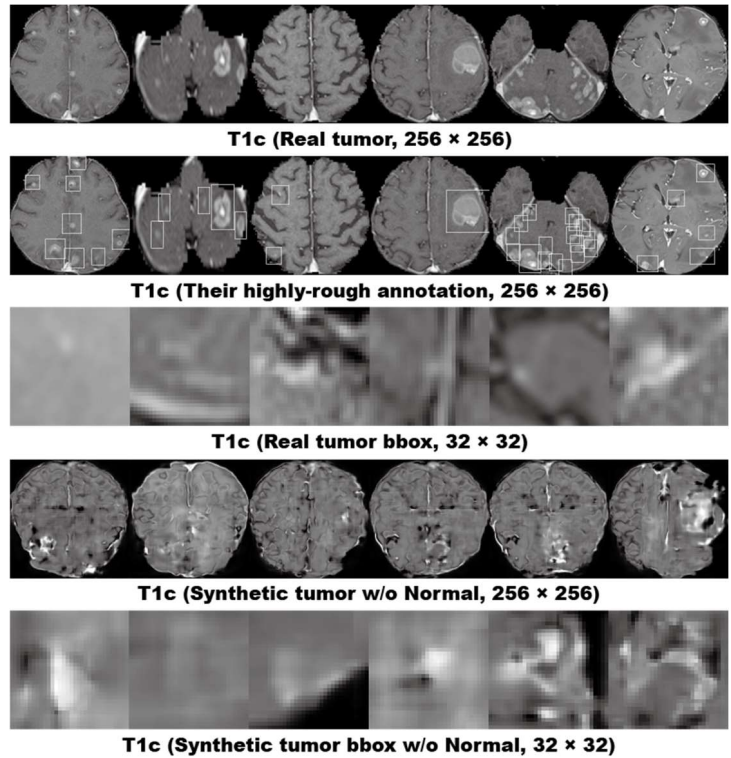

**Fig. 1** Example real/synthetic 256 × 256 MR images and resized 32 × 32 tumor bounding boxes.

## 4. Conclusion

Based on our experience and related work, we introduced GAN-based Medical Image Augmentation, along with the tricks to boost classification/object detection/segmentation performance using them. As an example, we also showed that our novel GAN can generate 256 × 256 MR images with brain metastases of random shape within bounding boxes, naturally at desired position/size, and achieve high sensitivity in tumor detection—even with small/fragmented training data from multiple MRI scanners and lazy annotation using highly-rough bounding boxes.

**Acknowledgement**

This research was supported by AMED under Grant Number JP18lk1010028. The authors appreciate Dr. Tomoyuki Noguchi for preparing the dataset at the Department of Radiology, National Center for Global Health and Medicine.

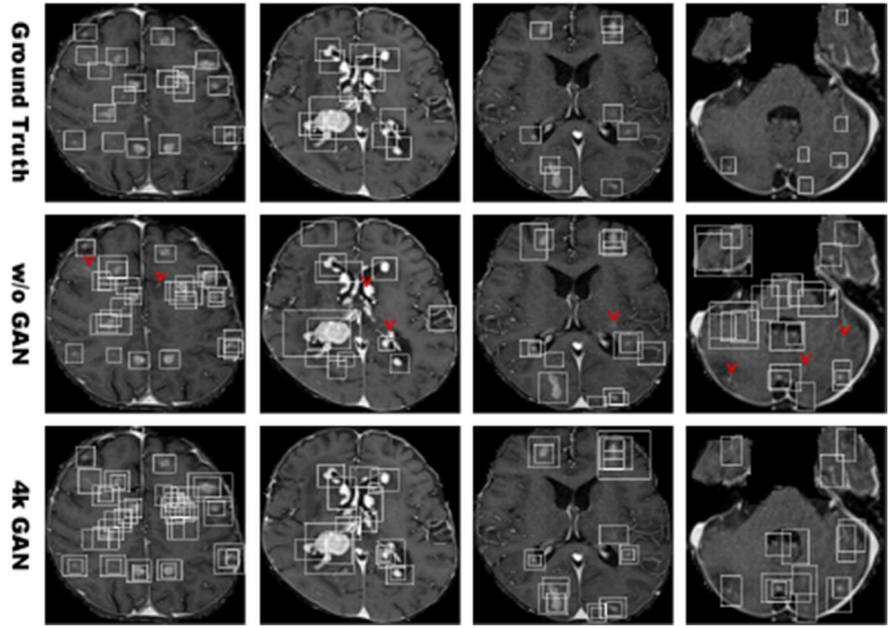

**Fig. 2** Example detection results obtained by the DA setups on four different images, compared against the ground truth: (a) Ground truth; (b) Without GAN-based DA; (c) With 4k GAN-based DA. Red V symbols indicate the brain metastases undetected without GAN-based DA, but detected with 4k GAN-based DA.

**Table 1** YOLOv3 brain metastases detection results with/without GAN-based DA using bounding boxes with 0.1% detection threshold.

|  |  | IoU ≥ 0.5 |  | IoU ≥ 0.25 |  |
|---|---|---|---|---|---|
|  | **mAP** | **Sensitivity** | **FPs per slice** | **Sensitivity** | **FPs per slice** |
| 2,813 real images | 0.51 | 0.67 | **4.11** | 0.83 | **3.59** |
| + 4,000 GAN-based DA | **0.54** | **0.77** | 7.64 | **0.91** | 7.18 |

robust lung segmentation. In Proc. International Conference on Medical Image Computing and Computer-Assisted Intervention (MICCAI): 732-740, 2018

## Biography

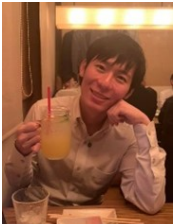

**Changhee Han**
　Ph.D. Student at Graduate School of Information Science and Technology, the University of Tokyo. He received a Master's Degree in Computer Science from the University of Tokyo. His research interests include Deep Learning for Medical Imaging.

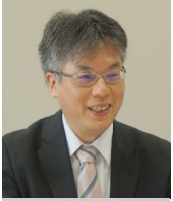

**Kohei Murao**
　Research Associate Professor at Research Center for Medical Bigdata, National Institute of Informatics. He received a Ph.D. Degree in Physics from Tohoku University. His research interests include Medical Image Analysis and Informatics.

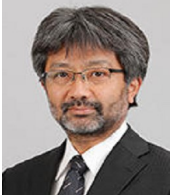

**Shin'ichi Satoh**
　Professor at Digital Content and Media Sciences Research Division, National Institute of Informatics. He received a Ph.D. Degree in Computer Science from the University of Tokyo. His research interests include Computer Vision and Pattern Media.

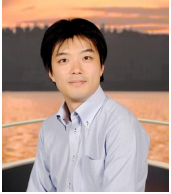

**Hideki Nakayama**
　Associate Professor at Graduate School of Information Science and Technology, the University of Tokyo. He received a Ph.D. Degree in Computer Science from the University of Tokyo. His research interests include Computer Vision and Deep Learning.